\documentclass{article}
\usepackage{ijcai26}

\usepackage{times}
\usepackage{soul}
\usepackage{url}
\usepackage[hidelinks]{hyperref}
\usepackage[utf8]{inputenc}
\usepackage[small]{caption}
\usepackage{graphicx}
\usepackage{amsmath}
\usepackage{amsthm}
\usepackage{booktabs}
\usepackage{algorithm}
\usepackage{algorithmic}
\usepackage[switch]{lineno}

\usepackage{amssymb}
\usepackage{stmaryrd}
\usepackage{latexsym}
\usepackage{array} 
\usepackage{adjustbox} 
\usepackage{multirow}  
\usepackage{subcaption} 
\usepackage{wrapfig} 

\title{PALCAS: A Priority-Aware Intelligent Lane Change Advisory System for Autonomous Vehicles using Federated Reinforcement Learning}

\author{
Yassine Ibork$^1$
\and
Nhat Ha Nguyen$^1$\and
Myounggyu Won$^{2}$\And
Lokesh Das$^1$\\
\affiliations
$^1$School of Computing, Wichita State University\\
$^2$Department of Computer Science, University of Memphis\\
\emails
\{yxibork, hxnguyen18\}@shockers.wichita.edu,
mwon@memphis.edu,
lokesh.das@wichita.edu
}

\begin{document}

\maketitle

\begin{abstract}
We present a priority-aware intelligent lane change advisory system based on multi-agent federated reinforcement learning, namely PALCAS, for autonomous vehicles (AVs). While existing lane-change approaches typically focus on single-agent systems or centralized multi-agent systems, we introduce a federated reinforcement learning-based multi-agent lane change system prioritizing lane changing based on vehicle destination urgency. PALCAS incorporates a novel priority-aware safe lane-change reward function to enable judicious lane-change decisions in both mandatory and discretionary scenarios. PALCAS leverages the parameterized deep Q-network (PDQN) algorithm to facilitate effective cooperation among agents, enabling both lateral and longitudinal motion controls of AVs. Extensive simulations conducted using the SUMO traffic simulator and Mosaic V2X communication framework demonstrate that PALCAS significantly improves traffic efficiency, driving safety, comfort, destination arrival rates, and merging success rates compared to baseline methods.
\end{abstract}

\section{Introduction}
Lane-change decision-making in urban highway driving is a challenging task, as it involves coordinating vehicles’ lateral and longitudinal motion controls in response to dynamically changing traffic patterns~\cite{nilsson2016lane,butakov2014personalized}. According to the NHTSA, 556,309 vehicles were involved in merging and lane-change–related crashes in the United States in 2023, resulting in 80,291 injuries and 838 fatalities~\cite{NHTSA2023}. Lane-change maneuvers contribute more than $10\%$ of traffic accidents in highways~\cite{wang2018reinforcement}, and $94\%$ of them occur due to human-driving errors~\cite{Singh2018CriticalReasons}. The NHTSA reported that human drivers make mistakes $42\%$ of the time due to improper perception and fail to carry out critical decisions $33\%$ of the time~\cite{Singh2018CriticalReasons}. Furthermore, improper lane changes account for $10\%$ of traffic congestion, significantly reducing overall traffic efficiency~\cite{ni2020multivehicle}.

Thanks to rapid advancements in communication technologies and artificial intelligence (AI), connected and autonomous vehicles (CAVs) are emerging as a promising solution for next-generation intelligent lane-change systems. 
Reinforcement learning (RL)-based lane-change methods allows CAVs to make adaptive and data-driven decisions in complex traffic situations~\cite{wang2025robust}. 
Single-agent lane change systems are common in the literature due to their ability to balance efficiency, safety of the ego vehicle, leading to a well-balanced lane change decision-making~\cite{yu2025hierarchical,chen2024human,zhuang2024hgrl,huang2023human}. Zhuang \emph{et al.} incorporated human preferences in training a PPO agent to acheive human-like driving in lane change scenario~\cite{zhuang2024hgrl}. Huang \emph{et al.} incorporated expert knowledge into the safety reward with human behavior filter to train a safe and efficiency agent in a mandatory lane-change scenario~\cite{huang2023human}. 
To ensure coordinations among agent and improve traffic safety and efficiency, researchers introduce multi-agent lane changes ~\cite{wang2024multi,bi2025mix,guo2024heuristic}. Wang \emph{et al.} proposed a safety-aware lane change policy that reduces collisions and lane change-induced traffic buildup~\cite{wang2024multi}, while Bi \emph{et al.} incorporated decision priority into the QCOMBO algorithm to promote strategic collaboration among the agents~\cite{bi2025mix}. Although, MARL highlights an immense potential in autonomous lane change decision making, the centralized MARL approaches may risk data insecurity and higher computational overhead.

Recent research has increasingly focused on federated learning (FL), particularly federated reinforcement learning (FedRL), for modeling multi-agent decision-making~\cite{fan2023blockchain,lu2025preference}. Unlike centralized RL approaches, FedRL distributes the training process across participating agents, while a central federated server is used solely for model aggregation. This distributed framework promotes data diversity and effective knowledge sharing among agents, enhances network efficiency, and reduces dependency on centralized systems. Nguyen \emph{et al.} introduced a generalizable supervised FL network, which enables agents to collaboratively learn to predict the steering angle using RGB road images~\cite{nguyen2022deep}. Du \emph{et al.} developed a personalized FL-based LSTM model that uses varying sequence lengths to predict highway lane changing in different behavior clusters~\cite{du2023driver}. Researchers further combine FL with RL and have shown promising results in collision avoidance and traffic signal control (TSC) tasks~\cite{fu2022selective,lu2025preference}. 
However, current FedRL studies in autonomous driving primarily address low-level motion control for collision avoidance and have yet to be explored in the context of integrated multi-agent lane-change decision-making~\cite{chellapandi2023federated}.

In this paper, we present \textbf{PALCAS}—a \emph{Priority-Aware Lane-Change Advisory System} for city highways with frequent on- and off-ramps—developed within a federated multi-agent reinforcement learning (Fed-MARL) framework. PALCAS partitions the highway into multiple clusters, each managed by a roadside unit (RSU) that coordinates the motion of connected and automated vehicles (CAVs) within its coverage zone. Through infrastructure-to-infrastructure (I2I) communication, RSUs share global traffic information to make coordinated, priority-aware decisions across the entire network. To the best of our knowledge, this is the first work that jointly addresses \emph{mandatory} and \emph{discretionary} lane-change behaviors using a novel priority-guided reward design integrated with a federated learning paradigm for large-scale cooperative decision-making. We formulate the priority-aware lane-change problem as a decentralized partially observable Markov decision process (Dec-POMDP) with a hybrid action space that captures both discrete lane-change maneuvers and continuous longitudinal control. The state space fuses microscopic (vehicle-level) and macroscopic (cluster-level) traffic features, enabling RSUs to reason about local and global dynamics simultaneously. Using parametrized deep Q-networks, PALCAS is trained and evaluated in a realistic V2X-enabled simulation environment built with SUMO and Eclipse MOSAIC. Comprehensive experiments demonstrate that PALCAS significantly enhances highway safety, comfort, and efficiency. The key contributions of this work are summarized as follows:

\begin{itemize}
    \item We propose \textbf{PALCAS}, a first-of-its-kind priority-aware multi-agent lane-change advisory framework built on federated reinforcement learning, enabling coordinated and privacy-preserving knowledge sharing among agents.
    \item We design a hierarchical state representation that integrates microscopic CAV and surrounding-traffic data with macroscopic cluster-level information from RSUs, enhancing situational awareness and decision quality in complex highway environments.
    \item We develop a priority-guided reward mechanism that unifies mandatory and discretionary lane-change behaviors by jointly modeling the feasibility and urgency of maneuvers under dynamic traffic conditions.
    \item We perform comprehensive simulations using realistic V2X communication setup across diverse traffic scenarios, demonstrating the proposed system’s scalability, safety, and real-world applicability.
\end{itemize}

\section{Design of PALCAS}
\subsection{Overview}

\begin{figure}[!htbp]
    \centering
    \includegraphics[width=\columnwidth]{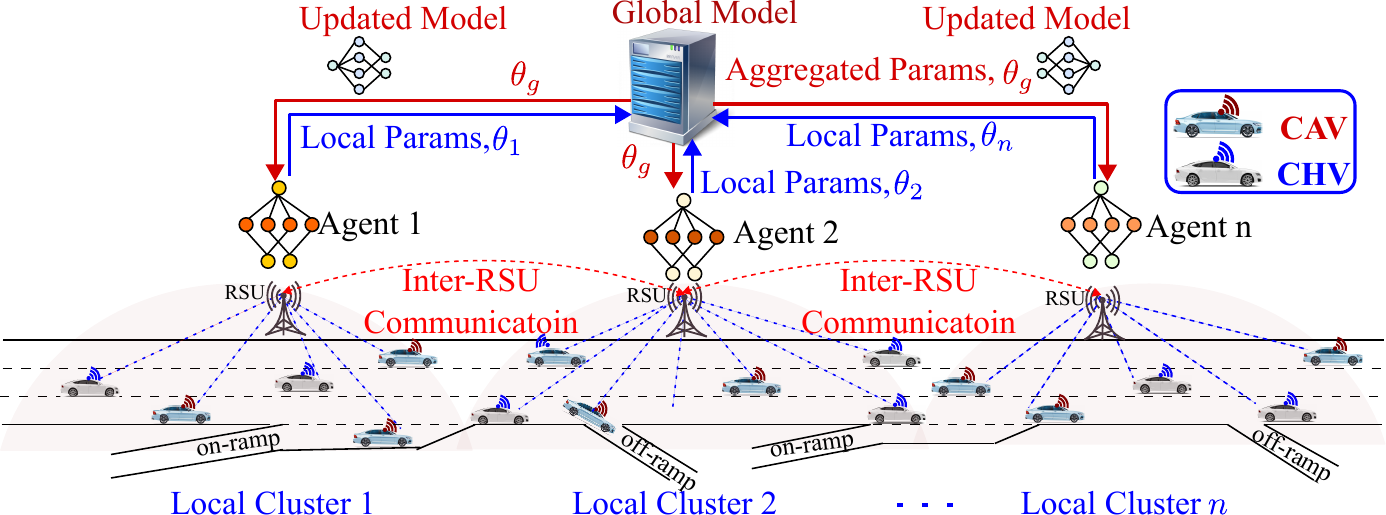}
    \caption{An operational overview of PALCAS.}
    \label{fig:system_overview}
\end{figure}

PALCAS is designed for urban highways characterized by frequent on- and off-ramps and high traffic volumes, as illustrated in Fig.~\ref{fig:system_overview}, where CAVs and connected and human-driven vehicles (CHVs) coexist. A long multi-lane highway is partitioned into multiple segments, with each segment falling within the coverage area of an RSU. Each RSU is designated to manage a specific highway segment, forming a local cluster comprising all CAVs within its communication range. Following common approaches in the literature~\cite{yadavalli2023rlpg}, RSUs function as agents that determine the actions of CAVs within their local coverage areas. In PALCAS, each CAV’s RSU membership adaptively switches as it moves from the coverage area of one RSU to another.
A unique aspect of PALCAS is its ability to fuse local and global traffic information: each RSU coordinates with others via I2I communication to make globally informed yet locally optimized motion-control decisions for CAVs using Fed-MARL. This integration enables scalable, cooperative control across dynamic highway environments. Particularly, each agent (\emph{i.e.}, RSU) independently trains a MARL model using local observations and periodically uploads its learned parameters to a central server. The server aggregates received parameters to update the global model and broadcasts the updated weights to all agents. Each agent then synchronizes its local model with the global weights and interacts with the environment accordingly. This federated learning setup is particularly advantageous in long-distance, dynamic, and heterogeneous traffic conditions, as it enables agents to generalize from unseen scenarios.

\subsection{Problem Formulation}
The proposed priority-aware lane-change advisory system is developed within a Fed-MARL framework, where the multi-agent lane-change problem is formulated as a Decentralized Partially Observable Markov Decision Process (Dec-POMDP). The Fed-MARL architecture leverages federated learning to train the model across distributed agents. The Dec-POMDP is defined by the tuple $\mathcal{P} = \langle \mathcal{K}, \mathcal{S}, \{\mathcal{A}_k, \mathcal{O}_k, \mathcal{R}_k\}_{k \in \mathcal{K}}, \mathcal{T}, \Omega, \gamma \rangle$, where $\mathcal{K}$ denotes the set of agents (\emph{i.e.}, RSUs) and $\mathcal{S}$ represents the global state of the highway environment. For each agent $k \in \mathcal{K}$, $\mathcal{A}_k$ is the action space corresponding to the lane-change advisories issued by RSU $k$ to vehicles in its controlled cluster. At time step $t$, each agent $k$ receives a local observation $o_k^t \in \mathcal{O}_k$. The joint observation and action spaces defined as $\mathcal{O} = \prod_{k \in \mathcal{K}} \mathcal{O}_k$ and $\mathcal{A} = \prod_{k \in \mathcal{K}} \mathcal{A}_k$, respectively. The state-transition function $\mathcal{T}: \mathcal{S} \times \mathcal{A} \times \mathcal{S} \to [0,1]$ defines the probability of transitioning to $s_{t+1}$ given $(s_t, a_t)$, while the observation function $\Omega: \mathcal{S} \times \mathcal{A} \times \mathcal{O} \to [0,1]$ specifies the likelihood of observing $o_t$ after executing $(s_t, a_t)$ and reaching $s_{t+1}$. The reward function $\{\mathcal{R}_k\}_{k \in \mathcal{K}}$ encodes safety, efficiency, comfort, and the \emph{proposed priority-aware lane-change reward} components. The overall learning objective is to maximize the cumulative discounted return across all agents: $\mathcal{G}(\pi) = \sum_{k \in \mathcal{K}} \mathbb{E}\!\left[\sum_{t=0}^{\infty} \gamma^{t}\, \mathcal{R}_k(s_t, a_t, s_{t+1})\right]$, where $\gamma \in (0,1)$ is the discount factor.

To model both discrete maneuver decisions and continuous control parameters, we define a hybrid action space for each CAV $u$ under RSU $k$ as $\mathcal{A}_{k,u}(s)=\bigl\{(q,c_q)\ \big|\ q\in\{1,\dots,Q\},\ c_q\in\mathcal{C}_q\bigr\},$
where $q$ indexes an action and $c_q$ is its associated continuous parameter. More specifically, the action space of agent $k$ (\emph{i.e.}, the $k^{\text{th}}$ RSU) consists all CAVs' actions,
$
\mathcal{A}_k(s)=\prod_{u\in U_k(s)} \mathcal{A}_{k,u}(s)$, where $U_k$ is a set of CAVs controlled by agent $k$ and
$\mathcal{A}(s)=\prod_{k\in\mathcal{K}}\mathcal{A}_k(s).$ We adopt PDQN~\cite{xiong2018parametrized} to handle the hybrid structure of the action space. At time step $t$, a parameter network $\mu_{\theta}$ outputs continuous parameters
$c_t=\mu_{\theta}(s_t)$ for each action, and a Q-network $Q_{\phi}$ evaluates
$Q_{\phi}(s_t,q,c_q)$ for all actions $q\in\{1,\dots,Q\}$. The action to be executed is determined through
$
q_t^{\ast}=\arg\max_{q} \, Q_{\phi}\bigl(s_t, q, c_q\bigr),
$
with the associated parameter $c_{q_t^{\ast}}$ taken from $c_t$. The RSU’s action $a_k^t$ is formed by stacking the selected per vehicle actions $\bigl(q_{k,u}^t, c_{q_{k,u}^t}^t\bigr)$ over $u\in U_k(s_t)$. We leverage FedRL to improve cross-agent learning. Each agent $k$ trains its own PDQN on trajectories collected within its cluster and periodically participates in model aggregation via FedAvg~\cite{mcmahan2017communication} by sharing local weights with the central server. PALCAS optimizes the global objective over all $\mathcal{K}$ RSUs as, $\min_{\theta\in{R^d}}F(\theta)=\sum_{k=1}^{\mathcal{K}}{n_{k}\over{n}}\mathcal{L}_{k}(\theta), \text{ with } n= \sum_{k=1}^{\mathcal{K}}n_k,$
where $\mathcal{L}_k$ is the local loss of agent $k$ and $n_k$ is the number of local samples collected by agent $k$ in the current round. After $\mathcal{E}$ local steps, the server aggregates the weights by a size-weighted average, $\theta^{r+1}=\sum_{k=1}^{\mathcal{K}}{n_{k}\over{{n}}}\mathcal{\theta}_{k}^{r}$, 
where $\theta_k^{r}$ denotes agent $k$’s weights at the end of $r$ communication rounds before aggregation.

\subsection{State Space\label{sec:state-space}}
As each RSU is responsible for managing CAVs within its coverage zone, the state space of agent $k$ is defined as the aggregate state of all CAVs under its control: $\mathcal{S}_t^k = \bigl\{ s_{\text{ego}}^i,\, s_{\text{sur}}^i,\, s_{\text{local}}^k,\, s_{\text{global}} \ \big|\ i \in U_k(t) \bigr\},$ where $U_k(t)$ denotes the set of CAVs managed by RSU $k$ at time $t$. More specifically, each RSU determines the optimal lane-change decision for $i^{\text{th}}$ CAV by leveraging both local and global information. The local information includes (1) the ego-vehicle state $s_{\text{ego}}^i$ which comprises lateral and longitudinal positions, velocity, acceleration, current lane, and distance to the highway exit, (2) the surrounding vehicle state $s_{\text{sur}}^i \in \mathbb{R}^{\mathcal{D} \times \mathcal{F}}$ representing information about $\mathcal{D}$ neighboring vehicles with $\mathcal{F}$ features each, such as relative distance to the $i^{\text{th}}$ CAV, velocity, acceleration, lane index, and distance to exit, and (3) the cluster-level state $s_{\text{local}}^k$ capturing macroscopic features including the average speed of the cluster and lane-wise traffic densities. The global state $s_{\text{global}}$ represents aggregate traffic statistics (e.g., average cluster densities) shared among RSUs via I2I communication. This hybrid state representation enables each agent to maintain traffic awareness beyond its local observation range.

\subsection{Action Space}
The lateral and longitudinal motion control of each CAV is jointly considered in the design of the action space. In other words, each RSU can issue either a lane-change or a longitudinal control command (acceleration or deceleration) to a CAV within its coverage zone. More specifically, agent $k$ selects an action $a_i \in \{a_0, a_1, a_2, a_3\}$ for the $i^{\text{th}}$ CAV, where $a_0$ denotes a lane change to the left, $a_1$ a lane change to the right, $a_2 \in [a_{\text{min}}, a_{\text{max}}]$ a continuous acceleration command, and $a_3$ maintaining the current cruising speed.

\subsection{Reward Function} 

Our multi-objective reward function for optimizing the federated-learning-based lane-change system is defined as follows.
\begin{equation}
\resizebox{.899\columnwidth}{!}{$
    \mathcal{R}_k^{(t)} =\sum_{i\in {S_{\text{CAVs}}}} \left( \iota r_e^{(t)} + \zeta r_s^{(t)} + \xi r_c^{(t)} + \lambda r_{lc}^{(t)} + \psi r_d^{(t)} \right)
    $}.
\end{equation}
\noindent  The reward function consists of a traffic-efficiency reward $r_e^{(t)}$, a safety reward $r_s^{(t)}$, and a driving-comfort reward $r_c^{(t)}$, along with a \textbf{novel priority-based lane-change reward} $r_{lc}^{(t)}$. In addition, a deadlock penalty reward $r_d^{(t)}$ is incorporated to encourage early merging of ramp vehicles. Here, $\iota$, $\zeta$, $\xi$, $\lambda$, and $\psi$ are weighting coefficients to prioritize one reward component over others. In our experiments, we set \(\zeta = 0.5\), \(\lambda = 0.4\), \(\psi = 0.1\), \(\iota = 0.05\), and \(\xi = 0.05\).

\noindent \textbf{Efficiency Reward:} The efficiency reward is designed to enhance both individual CAV performance and the overall traffic efficiency within each RSU-managed cluster. It is defined as a weighted combination of the ego-vehicle efficiency $r_{\text{ego}}^{(t)}$ and the cluster-level efficiency $r_{\text{cluster}}^{(t)}$:
\begin{equation}
    r_e^{(t)} = w_c \, r_{\text{cluster}}^{(t)} + w_e \, r_{\text{ego}}^{(t)}.
\end{equation}
$r_{\text{ego}}^{(t)}$ encourages the CAV to maintain a speed close to the maximum allowable limit $v_{\text{max}}^{\text{ego}}$, while penalizing excessive speeding (\emph{i.e.}, $v_{\text{ego}}^{(t)} > v_{\text{max}}^{\text{ego}}$). It is formulated as $ r_{\text{ego}}^{(t)} = 
    -(|(v^{(t)}_{\text{ego}}-v_{\text{max}}^{\text{ego}}|)/(v_{\text{max}}^{\text{ego}}-v_{\text{min}}^{\text{ego}})$.
Similarly, $r_{\text{cluster}}^{(t)}$ enables each RSU to maintain the average speed of its controlled zone close to the target maximum mean speed $\bar{v}_{\text{max}}$. It is expressed as $r_{\text{cluster}}^{(t)} = -(|\bar{v}_{\text{cluster}}^{(t)}-\bar{v}_{\text{max}}|)/(\bar{v}_{\text{max}}-\bar{v}_{\text{min}})$.

\noindent \textbf{Safety Reward:} The safety reward is designed based on the \emph{Responsibility-Sensitive Safety} (RSS) model~\cite{shalev2017formal}, which formally defines dynamic lateral and longitudinal safety thresholds for autonomous driving. The RSS model computes the minimum longitudinal safe distance that the $i^{\text{th}}$ CAV must maintain from its leading vehicle as
\begin{equation}
    \delta_{\text{long}}^{(t)} = \Big[v^{(t)}_i \rho + \tfrac{1}{2} a_{\text{max}} \rho^2 + \tfrac{(v^{(t)}_i+\rho a_{\text{max}})^2}{2a_{b,\text{min}}} - \tfrac{(v^{(t)}_{\text{lead}})^2}{2a_{b,\text{max}}}\Big]_{+},
    \label{eq:long-safe_distance}
\end{equation}
where $[X]_{+} := \max(X,0)$. Here, $\rho$ denotes the reaction time of the CAV to the leading vehicle's deceleration, which is set to $0.2\text{s}$~\cite{kim2021development}. $v^{(t)}_i$ and $v^{(t)}_{\text{lead}}$ are the longitudinal velocities of the ego and leading vehicles, respectively. $a_{\text{max}}$ represents the maximum acceleration a CAV may apply during the reaction time, while $a_{b,\text{max}}$ and $a_{b,\text{min}}$ denote the maximum braking of the leader and the minimum braking required by the ego vehicle to avoid collision. We set $a_{\text{max}} = 2.6\text{m/s}^2$ and $a_{b,\text{min}} = a_{b,\text{max}} = 4.5\text{m/s}^2$, respectively~\cite{SUMO2018}.

\noindent The RSS model also defines the minimum lateral safe distance between the ego vehicle and its four nearest adjacent vehicles $m \in \{1,2,3,4\}$ as
\begin{equation}
    \resizebox{.899\columnwidth}{!}{$
            \displaystyle
            \delta_{\text{lat}}^{(t)} = \mu + \Big[\tfrac{(v^{(t)}_i+v^{(t)}_{i,\rho})}{2}\rho + \tfrac{(v^{(t)}_{i,\rho})^2}{2a_{b,\text{lat}}}
        - \big(\tfrac{(v^{(t)}_m+v^{(t)}_{m,\rho})}{2}\rho - \tfrac{(v^{(t)}_{m,\rho})^2}{2a_{b,\text{lat}}}\big) \Big]_{+}
        $},
        \label{eq:lateral-safety}
\end{equation}%

\noindent where $\mu = 0.1\text{m}$ is the minimum clearance buffer~\cite{de2021universally}. $v_i$ and $v_m$ are the lateral velocities of the ego and adjacent vehicles, respectively. $a_{b,\text{lat}} = 1\text{m/s}^2$ denotes the minimum lateral braking deceleration during the response time $\rho$. The lateral response velocities are given by $v_{i,\rho}^{(t)} = v_i^{(t)} + \rho a_{\text{lat,max}}$ and $v_{m,\rho}^{(t)} = v_m^{(t)} + \rho a_{\text{lat,max}}$, where $a_{\text{lat,max}} = 1\text{m/s}^2$ is the maximum lateral acceleration.

\noindent At each time step, the CAV evaluates longitudinal and lateral safety rewards as
\begin{equation}
    r_{s,\text{long}}^{(t)} = \sum_{m \in \{\text{lead}, \text{follow}\}} \min\!\Big(\tfrac{d_{\text{long}, m}^{(t)} - \delta_{\text{long}, m}^{(t)}}{\delta_{\text{long}, m}^{(t)}}, 0\Big),
\end{equation}
\begin{equation}
    r_{s,\text{lat}}^{(t)} = \sum_{m=1}^{4} \min\!\Big(\tfrac{d_{\text{lat}, m}^{(t)} - \delta_{\text{lat}, m}^{(t)}}{\delta_{\text{lat}, m}^{(t)}}, 0\Big),
\end{equation}
where $d_{\text{long},m}^{(t)}$ and $d_{\text{lat},m}^{(t)}$ denote the actual longitudinal and lateral distances between the ego and surrounding vehicles, respectively. The final safety reward is computed as $r_s^{(t)} = r_{s,\text{long}}^{(t)} + r_{s,\text{lat}}^{(t)}.$

\noindent \textbf{Comfort Reward:} The comfort reward ensures that CAV's acceleration, $a_i^{(t)}$ is within the comfortable threshold $a_{\text{th}}$, which is set to $1.47\text{m/s}^2$ according to the literature~\cite{de2023standards}. When a CAV’s instantaneous acceleration exceeds this threshold, a penalty is applied. 
The comfort reward is expressed as $r_c^{(t)} = (a_{\text{th}} - |a_i^{(t)}|)/(|a_{\text{min}}| - a_{\text{max}})$, where $a_{\text{min}} = -4.5\text{m/s}^2$ and $a_{\text{max}} = 2.6\text{m/s}^2$.

\noindent \textbf{Priority-guided Lane-Change Reward:} 
The priority-based lane-change reward guides CAVs’ lane-change decisions according to their exit urgency and local traffic efficiency. In essence, this reward encourages CAVs with imminent exits to move toward the rightmost lane while directing those with later exits toward middle or left lanes. It comprises a destination-based urgency term $u_t w_t$ and a cluster-oriented staging penalty $p_{\text{stage}}^{(t)}$, as $r_{lc}^{(t)} = u_t w_t + p_{\text{stage}}^{(t)}.$
\noindent $u_t w_t$ promotes timely lane changes for vehicles approaching their exits. $u_t$ becomes increasingly negative when a CAV nears its exit but remains in a non-exit lane. To calculate the lane change urgency, it compares the time available to the exit, $\tau_{\text{tte},t}$ with the time needed to complete the remaining lane changes $\tau_{\text{need},t}$, using a feasibility factor $p_t \in [0,1]$:
\begin{equation}
    u_t = 
    \begin{cases}
        -p_t, & \hspace{-1.6cm}\text{if } d_t < d^{\text{th}} \text{ and } v_{\text{ego}}^{(t)} < v^{\text{th}},\\
        -\dfrac{1}{1+\exp{(\tau_{\text{tte},t}-\tau_{\text{need},t})}}, & \text{otherwise}.
    \end{cases}
\end{equation}
Here, $d^{\text{th}} = 50\text{m}$ and $v^{\text{th}} = 5\text{m/s}$ are empirically chosen thresholds to prevent congestion caused by vehicles that are both close to the exit ($d_t \to 0$) and moving slowly ($v_{\text{ego}}^{(t)} \to 0$).

Let $l_t$ be the ego's (\emph{i.e.,} CAV's) current lane index and $\mathcal{L}$ the total number of lanes; $l^*$ is the exit lane assumed to be the rightmost lane indexed by $1$. The remaining lane count is $n_t = |l_t - l^*| \in \{0, . . ., \mathcal{L}-1\}$. Let $d_t$ be the remaining longitudinal distance to the exit (in $\text{m}$); $v_{\text{ego}}^{(t)}$ is the ego speed (in $\text{m/s}$); $\tau_0$ is the nominal time for a single lane change in free flow ($2\text{s}$~\cite{williams2018anticipatory}) and $\epsilon \ll 1$ is a constant to avoid divide by zero situation. Finally, we compute $\tau_{\text{need},t} = n_t\tau_0 /(p_t+\epsilon)$ and $\tau_{\text{tte},t}=d_t/(v_{\text{ego}}^{(t)}+\epsilon)$.

We laterally project ego vehicle position in the target lane and define $p_t$ from the projected time-to-collision ($\widehat{\text{p-ttc}}$) in the target lane $l_j$, rating the feasibility of reaching the exit lane, $l^{*}$, safely and comfortably. It is worth noting that $p_t$ is precisely designed to gauge the time, $\tau_{\text{need},t}$ required to change lanes by taking into account safety constraints, discouraging lane changes if there is a unsafe situation (\emph{e.g.,} $\widehat{\text{p-ttc}}_{\text{min},t} \le \text{ttc}^{*}$). In particular, $p_t$ allows $\tau_{{need},t}$ to span smoothly from feasible safe lane changes in linear time ($\approx n_t\tau_0$) to infeasible lane changes (\emph{e.g.,} nearly $\infty$ time). let $\Delta t = \widehat{\text{p-ttc}}_{\text{min},t}-{\text{ttc}}^*$, where $\widehat{\text{p-ttc}}_{\text{min},t}$ is the minimum projected time-to-collision from CAV to the target leader or from the target follower to CAV, and $\text{ttc}^*$ is the time-to-collision threshold which is set to $1.5\text{s}$~\cite{das2019defining}, then $p_t = \sigma(\frac{\Delta t}{\sigma_{\text{ttc}}})$, where the temperature parameter, $\sigma_{ttc} \in (0,\;1)$ controls how sensitive $p_t$ is in terms of lane changes and $\sigma$ is the sigmoid function. Let $\widehat{\text{p-ttc}}_{\text{min},t} \ll {\text{ttc}}^*$ and $\sigma_{\text{ttc}}\approx 0.1$, then $p_t$ will be very small, resulting much higher $\tau_{\text{need},t}$, indicating that the current lane change is not feasible due to the potential collision.

Let $v_t^{(j)}$ be the speed of vehicles in the target lane $l_j$, $d_t^{(l_j)}$ be the longitudinal distance between vehicles, and $l_{\text{lead}}$ be the length of the leading vehicle. The projected time-to-collision for the ego vehicle with respect to the leader is:
\begin{equation}
\text{p-ttc}_{\text{ego}}^{(t)} = \begin{cases}
\frac{d_{t,\text{lead}}^{(j)}-l_{\text{lead}}}{v_{\text{ego}}^{(t)}-v_{t,\text{lead}}^{(j)}}, & \text{ if } v_{\text{ego}}^{(t)} > v_{t,\text{lead}}^{(j)}\\
+\infty, & \text{otherwise},
\end{cases}
\end{equation}
and the projected time-to-collision for the following vehicle with respect to the ego vehicle is:
\begin{equation}
\text{p-ttc}_{\text{follow}}^{(t)} = \begin{cases}
\frac{d_{t,\text{ego}}^{(j)}-l_{\text{ego}}}{v_{\text{t, follow}}^{(j)}-v_{\text{ego}}^{(t)}}, & \text{ if } v_{\text{t, follow}}^{(j)} > v_{\text{ego}}^{(t)}\\
+\infty, & \text{otherwise}.
\end{cases}
\end{equation}
Finally, the minimum projected $\text{ttc}$ is defined as $\widehat{\text{p-ttc}}_{\text{min},t} = \text{min}(\text{p-ttc}_{\text{ego}}^{(t)}, \text{p-ttc}_{\text{follow}}^{(t)})$.\\
We scale the urgency $u_t$ by a dynamic weight $w_t$ which is designed by taking into account the \emph{exit proximity} and remaining \emph{lane-change (LC) factor} of CAVs, as defined, eq.~(\ref{eq:scaling-factor}). The \emph{exit proximity} is represented as a function of distance remaining to exit, $d_t$, and the remaining \emph{LC factor} is defined in terms of the number of lane changes required to reach $l^*$.
\begin{equation}
w_t = \underbrace{2*\Big(1-\sigma(\frac{d_t}{1000})\Big)}_{\text{Exit Proximity}} \times\underbrace{\Big(\frac{n_t}{\mathcal{L}-1}\Big)}_{\text{LC Factor}}.
\label{eq:scaling-factor}
\end{equation}
$w_t$ weighs $u_t$ with a higher value if the CAV is very close to the exit, but needs multiple lane changes to be in the exit lane. In contrast, $w_t$ proportionately neutralizes $u_t$ if the CAV is far from the exit and (or) already in the exit lane; thereby emphasizing more on the staging penalty, $p_{\text{stage}}^{(t)}$. $p_{\text{stage}}^{(t)}$ discourages a CAV occupying $l^*$ if it is too far from the exit. To incentivize this behavior,
$p_{\text{stage}}^{(t)}$ is defined as a product of the \emph{ early staging penalty} and \emph{exit lane occupancy}, which is an additive inverse of the LC factor, \emph{i.e.,}
\begin{equation}
p_{\text{stage}}^{(t)} = \underbrace{-2 * \Big(\sigma(\frac{d_t}{1000})-0.5\Big)}_{\text{Early Staging Penalty}}\times \underbrace{\Big(1-\frac{n_t}{\mathcal{L}-1}\Big)}_{\text{Inverse LC Factor}}.
\label{eq:staging-penalty}
\end{equation}
Specifically, $p_{\text{stage}}^{(t)}$ imposes a strong negative penalty for occupying the exit lane too early, while fading away the penalty when the exit is imminent or the ego remains in non-exit lanes.

\textbf{Deadlock Penalty:} The deadlock penalty reward is designed for ramp vehicles to prevent them from becoming stuck at the end of the acceleration lane. A CAV incurs a higher penalty, as defined in eq.~(\ref{eq:deadlock-penalty}), when it remains on the acceleration lane for an extended period:
\begin{equation}
    r_d^{(t)} =
    \begin{cases}
        - \exp{\!\Big(\dfrac{-(x - \text{len}_{\text{accel}})^2}{\beta \, \text{len}_{\text{accel}}}\Big)}, & \hspace{-0.1cm}\text{if on accel. lane}\\
        0, & \text{otherwise},
    \end{cases}
    \label{eq:deadlock-penalty}
\end{equation}
where $x$ denotes the vehicle’s current position along the acceleration lane, $\text{len}_{\text{accel}}$ is the total length of the lane, and $\beta$ is a scaling factor set to $10$~\cite{liu2022autonomous}.

\section{Simulation Results}
\subsection{Simulation Setup}
We employed SUMO~\cite{lopez2018microscopic} and Eclipse MOSAIC~\cite{schrab2022tits} simulators to develop PALCAS. SUMO was used to construct the road network and model microscopic vehicle dynamics, whereas Eclipse MOSAIC was utilized to simulate V2X communication. The Fed-MARL framework of PALCAS was implemented in PyTorch. The experimental setup considered a $2.4\text{km}$ five-lane city highway with multiple on- and off-ramps. The highway was divided into three segments, each controlled by an RSU managing a designated ``cluster'' of vehicles. Each cluster included one on-ramp, one off-ramp, and spanned a total length of $0.8\text{km}$. The mainline and on-ramp traffic flows were set to $3200$ and $600\;\text{veh/h/lane}$, respectively, with $60\%$ of the vehicles being CAVs. A warm-up zone of $100\text{m}$ was used to randomly insert vehicles into the simulation, with maximum speeds following a Bernoulli distribution. To enable fine-grained lane-change maneuvers, a lateral resolution of $3.2\text{m}$ was adopted based on the SL2015 model~\cite{semrau2016simulation}. Each CAV was allowed up to $2\text{s}$ to complete a lane-change maneuver~\cite{toledo2007modeling,ding2021safe}; otherwise, the attempt was aborted. Vehicles involved in collisions were removed from the simulation to maintain continuous traffic flow. Hyperparameter tuning was performed using a grid search to determine the optimal configuration. The PALCAS models were trained using batch normalization and the AdamW optimizer with a learning rate of $1e^{-3}$, weight decay of $1e^{-2}$, and ReLU activation function. The simulation parameters are summarized in Table~\ref{tab:simparams}.

\begin{table}
    \centering
    \begin{tabular}{lr}
        \toprule
        \multicolumn{2}{c}{\textbf{Environment Parameters}} \\
        \midrule
        Highway Speed Limit $v_{lim}$ & $33.528 \,\text{m/s}$\\
        Simulation Time Step $\Delta t$ & $0.1 \,\text{s}$ \\
        Time Gap Offset $\Delta t$ & $1 \,\text{s}$ \\
        Reaction Time & $0.1 \,\text{s}$ \\
        Max Lateral Accel./Decel. & $1 \,\text{m/s}^2$\\
        \bottomrule
        \multicolumn{2}{c}{\textbf{Reinforcement Learning Parameters}}\\ \midrule
        State Space, Action Space                  & $45,\;4$ \\
        Param. Action Size                         & $1$ (Acceleration) \\
        Hidden Layers                              & $[256,\,512,\,256]$ \\
        Loss Function                              & Huber Loss \\
        Federated Learning Updates, $\mathcal{E}$                 & $2500$ gradient steps \\
        Exploration $(\epsilon_{\text{init}},\,\epsilon_{\text{final}},\,\epsilon_{\text{decay}})$ & $(1.0,\,0.02,\,0.999985)$ \\
        Batch Size \& Dropout ($b$, $p$) & ($256, 0.1$) \\
        Discount Factor $\gamma$ & $0.995$ \\
        Replay Buffer Size $\mathcal{B}$ & $100{,}000$ \\
        Target Update Frequency $C$ & $15{,}000$ steps \\
        \bottomrule
    \end{tabular}
    \caption{Hyper/Params for PALCAS.}
    \label{tab:simparams}
\end{table}
We evaluated PALCAS using five key metrics: (i) collision rate (CR), (ii) destination success rate (DSR), (iii) merging success rate (MSR), (iv) comfort, and (v) efficiency. We compared PALCAS with two baselines: (1) Centralized Approach - A single agent observing the entire highway and controlling CAVs from all clusters, hereafter ``Baseline-1'', and (2) PALCAS without federated learning - Individual cluster trains its policy without data and knowledge sharing, hereafter ``Baseline-2'' by varying CAV's penetration rates (PRs). The key results are summarized in Table~\ref{tab:performance-over-pr}. It is worth noting that the proposed baselines are not derived from existing studies, as lane-change systems based on federated learning remain largely unexplored in the current literature.
\begin{table*}[!htbp]
  \centering
  \setlength{\tabcolsep}{2pt}%
  \renewcommand{\arraystretch}{1.6}%
  \begin{adjustbox}{max width=\textwidth}
  \begin{tabular}{@{}l*{12}{c}@{}}
    \toprule
    \multirow{2}{*}{PR (\%)} &
      \multicolumn{4}{c}{PALCAS} &
      \multicolumn{4}{c}{Baseline-1} &
      \multicolumn{4}{c}{Baseline-2} \\
    \cmidrule(lr){2-5}\cmidrule(lr){6-9}\cmidrule(lr){10-13}
      & Efficiency$\uparrow$ & CR (\%)$\downarrow$ & DSR (\%) $\uparrow$& MSR (\%) $\uparrow$ &
        Efficiency$\uparrow$ & CR (\%) $\downarrow$& DSR (\%) $\uparrow$& MSR (\%) $\uparrow$&
        Efficiency$\uparrow$ & CR (\%) $\downarrow$& DSR (\%) $\uparrow$ & MSR (\%) $\uparrow$ \\
    \midrule
     5  & $\mathbf{26.97 \pm 0.23}$ & $5.63 \pm 5.89$ & $\mathbf{83.18 \pm 5.47}$ & $87.19 \pm 5.93$ &
          $26.87 \pm 0.27$ & $\mathbf{4.80 \pm 6.10}$ & $80.26 \pm 4.88$ & $\mathbf{87.75 \pm 6.30}$ &
          $26.81 \pm 0.30$ & $8.00 \pm 4.21$ & $79.93 \pm 5.78$ & $87.22 \pm 5.36$ \\
    10  & $27.14 \pm 0.30$ & $5.70 \pm 4.42$ & $\mathbf{84.04 \pm 3.94}$ & $\mathbf{87.41 \pm 4.10}$ &
          $\mathbf{27.29 \pm 0.18}$ & $\mathbf{3.22 \pm 2.93}$ & $78.11 \pm 6.14$ & $86.97 \pm 4.41$ &
          $27.20 \pm 0.19$ & $8.31 \pm 5.30$ & $77.26 \pm 4.92$ & $85.46 \pm 5.22$ \\
    20  & $\mathbf{27.70 \pm 0.17}$ & $4.99 \pm 3.26$ & $\mathbf{89.07 \pm 4.53}$ & $89.75 \pm 4.70$ &
          $27.69 \pm 0.15$ & $\mathbf{4.73 \pm 2.18}$ & $76.66 \pm 5.72$ & $89.01 \pm 3.84$ &
          $27.63 \pm 0.19$ & $7.51 \pm 2.18$ & $76.27 \pm 7.26$ & $\mathbf{89.78 \pm 3.56}$ \\
    40  & $\mathbf{28.75 \pm 0.34}$ & $3.70 \pm 2.38$ & $\mathbf{90.15 \pm 3.43}$ & $92.31 \pm 3.27$ &
          $28.61 \pm 0.23$ & $\mathbf{3.04 \pm 1.27}$ & $62.19 \pm 5.22$ & $\mathbf{92.78 \pm 4.31}$ &
          $28.10 \pm 0.39$ & $8.12 \pm 2.14$ & $58.92 \pm 6.75$ & $91.79 \pm 3.71$ \\
    60  & $\mathbf{30.03 \pm 0.34}$ & $\mathbf{2.45 \pm 0.92}$ & $\mathbf{93.97 \pm 3.56}$ & $93.33 \pm 5.09$ &
          $29.12 \pm 0.31$ & $3.05 \pm 0.99$ & $40.34 \pm 8.84$ & $94.04 \pm 4.42$ &
          $28.10 \pm 0.33$ & $10.00 \pm 2.03$ & $43.93 \pm 9.70$ & $\mathbf{94.29 \pm 5.03}$ \\
    \bottomrule
  \end{tabular}
  \end{adjustbox}
  \caption{Performance of PALCAS and Baselines.}
  \label{tab:performance-over-pr}
\end{table*}
\subsection{Traffic Efficiency}
\begin{figure}[!htbp]
    \centering
    \includegraphics[width=\columnwidth]{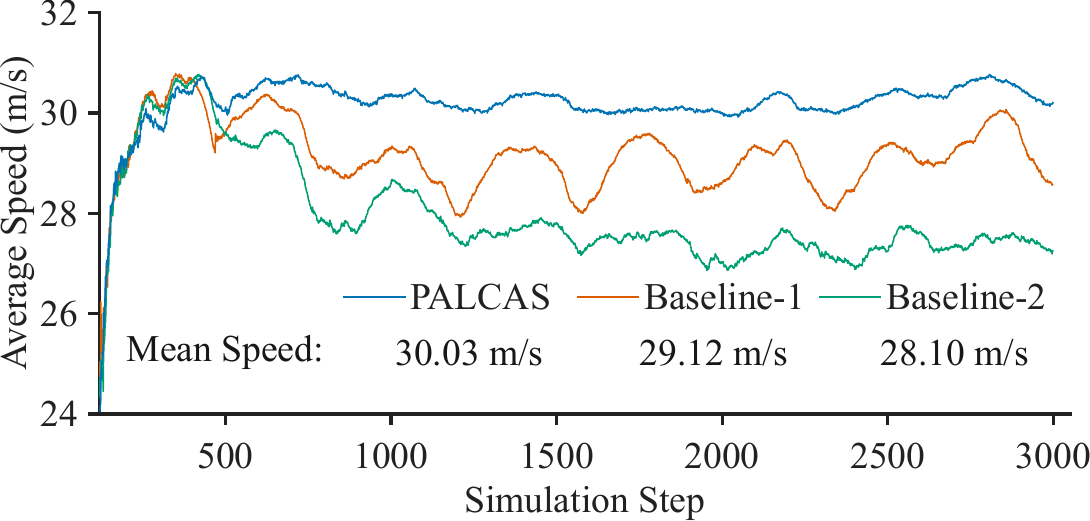}
    \caption{Changes in average speed over time.}
    \label{fig:speed-changes-over-time}
\end{figure}

\begin{figure}[!htbp]
  \centering
  \begin{subfigure}[b]{0.33\linewidth}
    \includegraphics[width=0.98\linewidth]{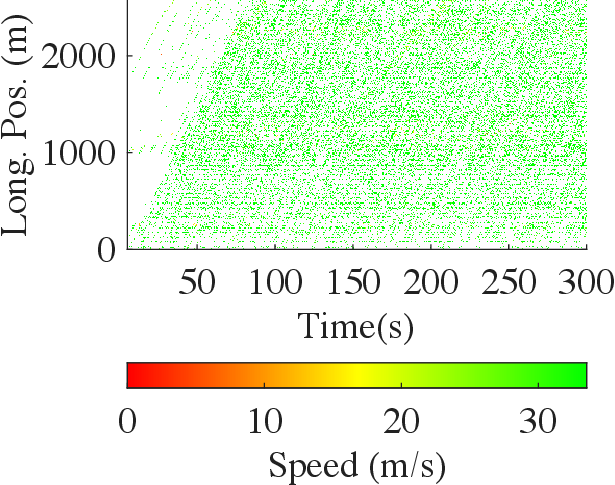}
    \caption{PALCAS}
    \label{fig:proposed}
  \end{subfigure}\hfill%
  \begin{subfigure}[b]{0.33\linewidth}
    \includegraphics[width=0.98\linewidth]{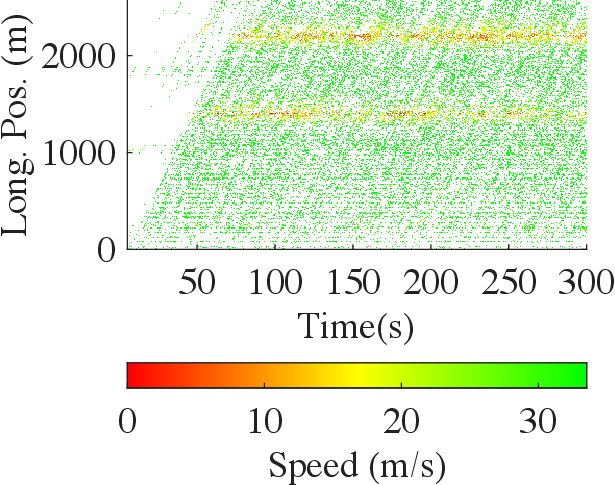}
    \caption{Baseline-1}
    \label{fig:baseline1}
  \end{subfigure}\hfill%
  \begin{subfigure}[b]{0.33
  \linewidth}
    \includegraphics[width=0.98\linewidth]{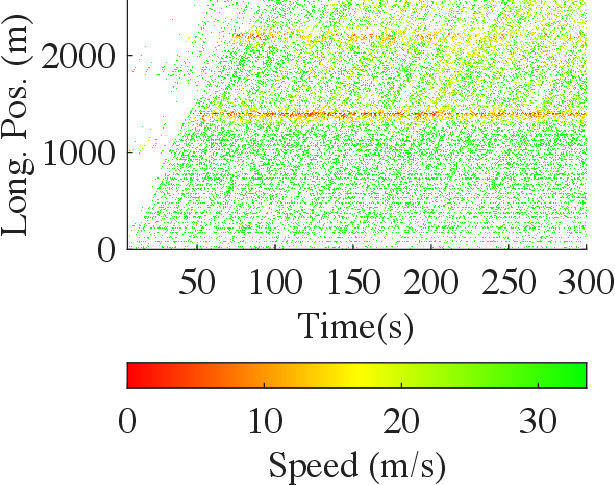}
    \caption{Baseline-2}
    \label{fig:baseline2}
  \end{subfigure}
  \caption{Space-time diagrams of average speed.}
  \label{fig:space-time-diagram}
\end{figure}

PALCAS evaluates traffic efficiency using the average speed of vehicles across all clusters. Fig.~\ref{fig:speed-changes-over-time} illustrates the average speed trends for all approaches. At the beginning of the simulation, all methods exhibit comparable performance since the traffic density is low and vehicles enter the network at maximum speed. As congestion builds up, however, the advantages of federated knowledge sharing and inter-RSU communication become evident. By leveraging global information beyond each CAV’s local perception range, PALCAS achieves superior cooperative lane-changing behavior, resulting in higher overall traffic efficiency. 
In particular, PALCAS maintains $3.12\%$ and $6.86\%$ higher average speeds than Baseline-1 and Baseline-2, respectively, throughout the entire simulation. The space–time diagram in Fig.~\ref{fig:space-time-diagram} further demonstrates PALCAS’s ability to alleviate intermittent traffic congestion caused by merging and exiting vehicles, underscoring its effectiveness in prioritizing lane changes under dynamically varying traffic conditions.
We further evaluate the impact of CAV penetration rate (PR) on traffic efficiency, with results summarized in Table~\ref{tab:performance-over-pr}. PALCAS consistently enhances traffic efficiency as the proportion of CAVs increases, raising the average speed from $26.97\text{m/s}$ at $5\%$ PR to $30.03\text{m/s}$, corresponding to an $11.34\%$ relative improvement. This gain stems from PALCAS’s architecture, which enables each RSU to make localized decisions within its cluster while collaboratively learning global traffic patterns through federated learning.

\subsection{Traffic Safety} 

We evaluate the safety performance of PALCAS using the collision rate, defined as the fraction of CAVs involved in collisions during the simulation. The results, summarized in Table~\ref{tab:performance-over-pr}, demonstrate that fewer CAVs experience collisions when controlled by PALCAS. As the CAV penetration rate (PR) increases, PALCAS consistently achieves a monotonic reduction in collision rate. Although Baseline-1 exhibits slightly better safety at lower PRs, PALCAS surpasses both baselines at higher penetration levels. Specifically, PALCAS reduces the collision rate by $19.67\%$ and $75.5\%$ compared to Baseline-1 and Baseline-2, respectively, at a PR of $60\%$. The centralized nature of Baseline-1 limits its ability to effectively coordinate decisions as the number of CAVs increases, while Baseline-2 performs the worst due to its lack of global knowledge sharing and inter-cluster coordination.

\subsection{Driving Comfort}

\begin{figure}[!htbp]
    \centering
    \begin{subfigure}{0.49\columnwidth}
        \includegraphics[width=0.98\textwidth]{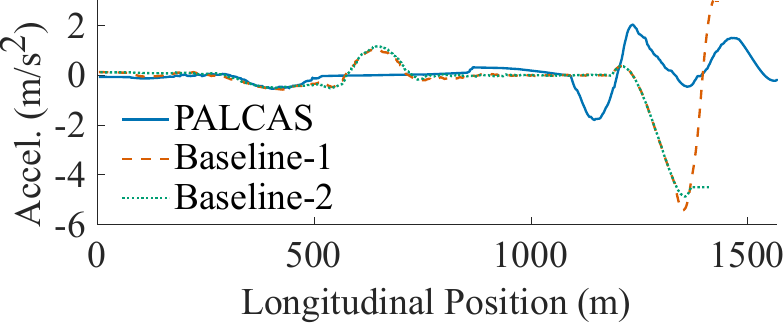}
        \caption{Route One }
        \label{fig:accel-profile-route-1}
    \end{subfigure}\hfill
    \begin{subfigure}{0.49\columnwidth}
        \includegraphics[width=0.98\textwidth]{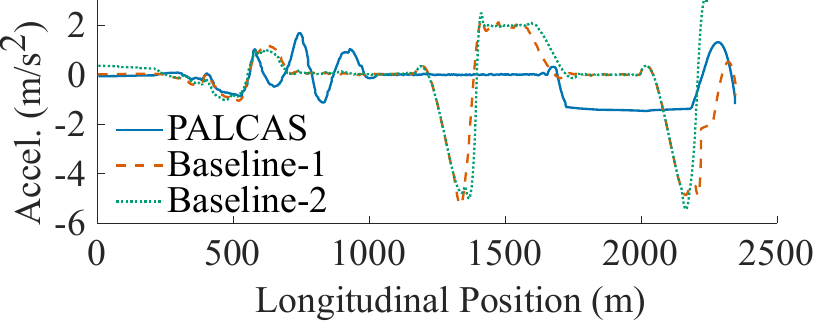}
        \caption{Route Two}
        \label{fig:accel-profile-route-2}
    \end{subfigure}\hfill
    \begin{subfigure}{0.49\columnwidth}
        \includegraphics[width=0.98\textwidth]{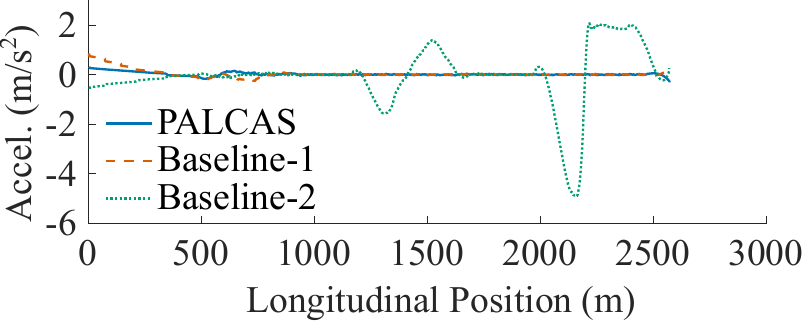}
        \caption{Route Three}
        \label{fig:accel-profile-route-3}
    \end{subfigure}\hfill
    \begin{subfigure}{0.49\columnwidth}
        \includegraphics[width=0.98\textwidth]{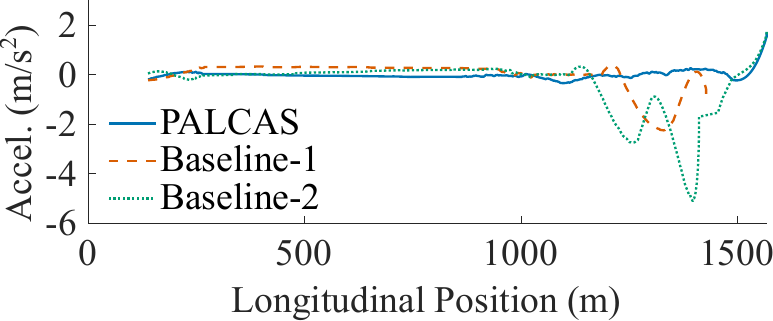}
        \caption{Route Four}
        \label{fig:accel-profile-route-4}
    \end{subfigure}\hfill
    \caption{Acceleration trajectories for CAVs on different routes. Routes are defined as (a) \textbf{Route One:} CAVs enter the mainline and exit to the cluster $2$, (b) \textbf{Route Two:} CAVs enter the mainline and exit to the cluster $3$, (c) \textbf{Route Three:} vehicles enter the mainline and continue to the end of the highway without exiting, (d) \textbf{Route Four:} vehicles take cluster-$1$'s ramp and exit to the cluster $2$.} 
    \label{fig:driving-comfort}
\end{figure}

To assess driving comfort, PALCAS analyzes the acceleration trajectories of vehicles across different routes, as shown in Fig.~\ref{fig:driving-comfort}. Each plot illustrates the acceleration profile of the same CAV under different control methods for a fair comparison. PALCAS produces smoother acceleration patterns than both baselines, maintaining acceleration within the comfortable range of $1.47\text{m/s}^2$~\cite{de2023standards}. In contrast, Baseline-1 and Baseline-2 exhibit larger fluctuations, with acceleration values mostly spanning $[-4.5,\,2.6]\text{m/s}^2$, whereas PALCAS keeps its acceleration tightly bounded within $[-1,\,1]\text{m/s}^2$.

\subsection{Success Rate}

\begin{figure}[!htbp]
    \centering
    \begin{subfigure}{0.49\columnwidth}
        \includegraphics[width=0.98\textwidth]{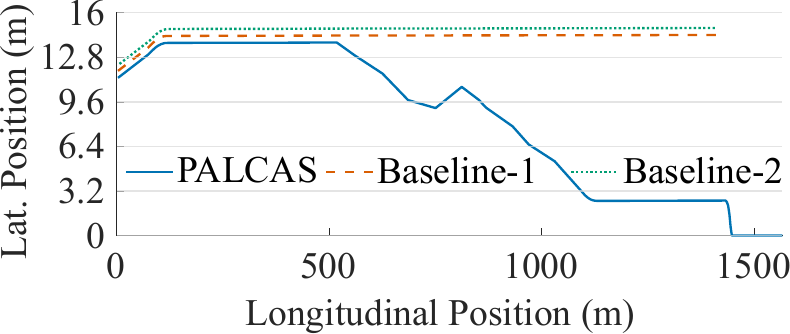}
        \caption{Route One}
        \label{fig:lane-change-route-1}
    \end{subfigure}\hfill
    \begin{subfigure}{0.49\columnwidth}
        \includegraphics[width=0.98\textwidth]{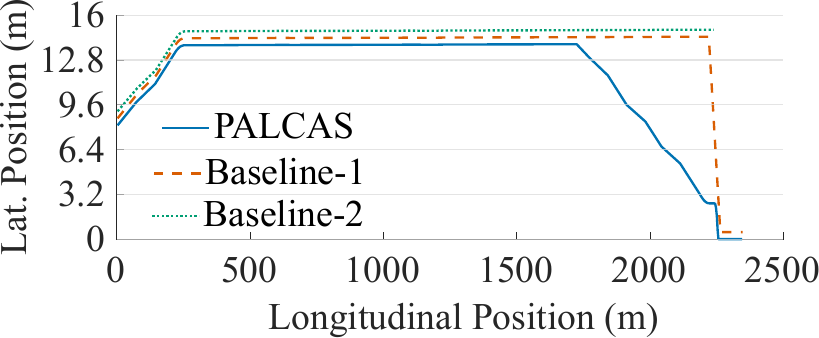}
        \caption{Route Two}
        \label{fig:lane-change-route-2}
    \end{subfigure}\hfill
    \begin{subfigure}{0.49\columnwidth}
        \includegraphics[width=0.98\textwidth]{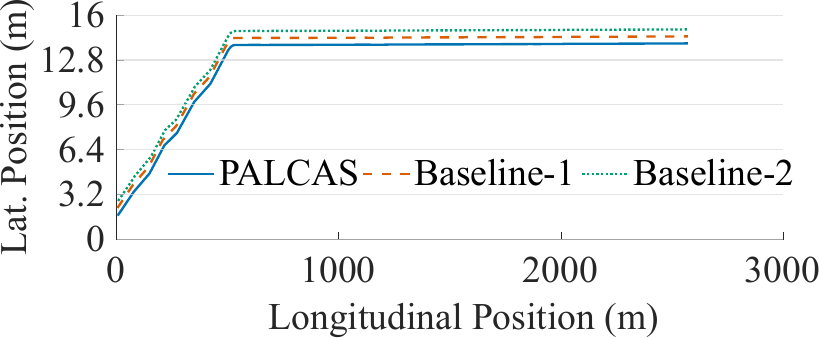}
        \caption{Route Three}
        \label{fig:lane-change-route-3}
    \end{subfigure}\hfill
    \begin{subfigure}{0.49\columnwidth}
        \includegraphics[width=0.98\textwidth]{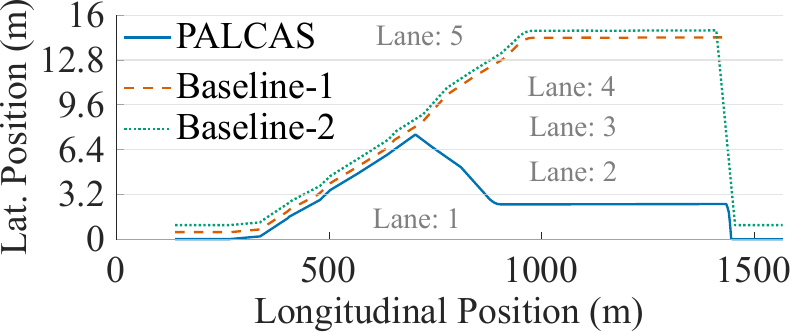}
        \caption{Route Four}
        \label{fig:lane-change-route-4}
    \end{subfigure}\hfill
    \caption{Lane-change trajectories of CAVs on routes. Baseline-1 and Baseline-2 are offset relatively by $0.5\text{m}$ and $1\text{m}$ to avoid overlapping with PALCAS.}
    \label{fig:lane-change-trajectory}
\end{figure}

Beyond traffic efficiency, safety, and comfort, we assess PALCAS’s generalization capability using MSR and DSR. MSR measures the fraction of merging CAVs that successfully enter the mainline without colliding or stopping at the end of the acceleration lane, while DSR represents the proportion of CAVs that successfully exit the highway to continue their routes. Table~\ref{tab:performance-over-pr} summarizes the MSR and DSR results for all methods. For DSR, PALCAS notably outperforms Baseline-1 and Baseline-2 by up to $3.63\%$ and $4.06\%$, respectively, even at a low penetration rate (PR) of $5\%$, and continues to achieve higher DSRs across all PR levels. At $60\%$ PR, PALCAS exceeds Baseline-1 and Baseline-2 by $132.94\%$ and $113.9\%$, respectively. These results are further supported by the lane-change trajectories illustrated in Fig.~\ref{fig:lane-change-trajectory}. PALCAS effectively prioritizes lane changes by balancing two complementary reward components: the \emph{staging term}, which promotes remaining in the fastest lane (lane~5) when the vehicle is far from its exit, and the \emph{urgency term}, which encourages timely transitions toward the rightmost lane (lane~1) as the exit approaches. 
As shown in Fig.~\ref{fig:lane-change-route-1} (see Fig.~\ref{fig:driving-comfort} for route definitions), a CAV controlled by PALCAS initially travels in the fastest lane (lane~5) when its exit is distant, then smoothly transitions toward the right, reaching the exit lane (lane~1) approximately $400\text{m}$ before the exit point. In contrast, the CAVs controlled by the baselines remain in the leftmost lane for too long and fail to execute timely lane changes, resulting in missed exits. A similar trend is observed in route~3 (Fig.~\ref{fig:lane-change-route-3}), where vehicles continue along the highway without exiting. The CAV using PALCAS correctly identifies that an exit is unnecessary and maintains its position in the leftmost lane, avoiding redundant lane changes that could otherwise disrupt traffic flow.

We further evaluate the performance of PALCAS and the baselines in terms of MSR. As presented in Table~\ref{tab:performance-over-pr}, PALCAS achieves a comparable MSR to both baselines, exceeding $93\%$ at a $60\%$ PR. Interestingly, Baseline-2 slightly outperforms PALCAS and Baseline-1 in MSR. This outcome can be attributed to the fact that, in Baseline-2, each RSU learns its policy independently, emphasizing localized optimization. Since merging is inherently a local maneuver within a specific cluster, each RSU in Baseline-2 prioritizes short-term local performance, thereby improving MSR. However, this myopic learning behavior limits Baseline-2’s ability to coordinate global maneuvers effectively, as reflected in its inferior performance on other metrics.

\subsection{Ablation Study}
 
Since the priority-guided lane-change reward is a key innovation of this study, we conducted an ablation experiment to evaluate its contribution to PALCAS’s overall performance. Specifically, we developed a variant of PALCAS excluding the priority-guided reward component and evaluated it across different PRs. The results, summarized in Table~\ref{tab:ablation-study}, reveal that the absence of this reward leads to a substantial performance degradation. In particular, without the priority-guided reward, PALCAS shows minimal sensitivity to increasing PRs—improving efficiency by only $4.77\%$ and reducing collision rate by $26.82\%$ as PR increases from $5\%$ to $60\%$. Moreover, the system struggles to coordinate multiple vehicles effectively, resulting in the lowest success rates across all metrics. These findings confirm that the priority-guided reward plays a critical role in enabling PALCAS to manage and coordinate CAVs safely and efficiently. It not only facilitates smooth and timely lane changes but also enhances merging behavior and overall traffic stability. In summary, the priority-guided reward is essential for achieving the cooperative intelligence required to handle complex highway scenarios with multiple on- and off-ramps.
\begin{table}
    \centering
    \begin{adjustbox}{max width=\columnwidth}
    \begin{tabular}{ccccc}
        \toprule
        PR (\%) & Efficiency$\uparrow$ & CR (\%) $\downarrow$& DSR (\%) $\uparrow$ & MSR (\%) $\uparrow$\\
        \midrule
        5  & 25.751 $\pm$ 0.715 & 35.97 $\pm$ 9.65 & 79.24 $\pm$ 5.86 & 80.86 $\pm$ 7.53 \\
        10 & 25.750 $\pm$ 0.548 & 35.21 $\pm$ 8.68 & 81.16 $\pm$ 6.24 & 77.38 $\pm$ 7.93 \\
        20 & 25.723 $\pm$ 0.424 & 34.79 $\pm$ 4.52 & 77.37 $\pm$ 8.43 & 70.25 $\pm$ 5.82 \\
        40 & 26.559 $\pm$ 0.301 & 27.80 $\pm$ 3.70 & 69.06 $\pm$ 10.15 & 58.95 $\pm$ 5.88 \\
        60 & 26.988 $\pm$ 0.340 & 26.32 $\pm$ 2.45 & 60.34 $\pm$ 12.09 & 49.54 $\pm$ 7.87 \\
        \bottomrule
    \end{tabular}
    \end{adjustbox}%

    \caption{PALCAS without priority-guided reward component.}
    \label{tab:ablation-study}
\end{table}

\subsection{Computational Overhead}
Federated learning adds computational complexity due to uploading (downloading) parameters to (from) the central server~\cite{fu2022selective,fu2024secure}. In addition to that,
\begin{wrapfigure}{r}{0.375\columnwidth}
    \centering
    \includegraphics[width=0.9\linewidth]{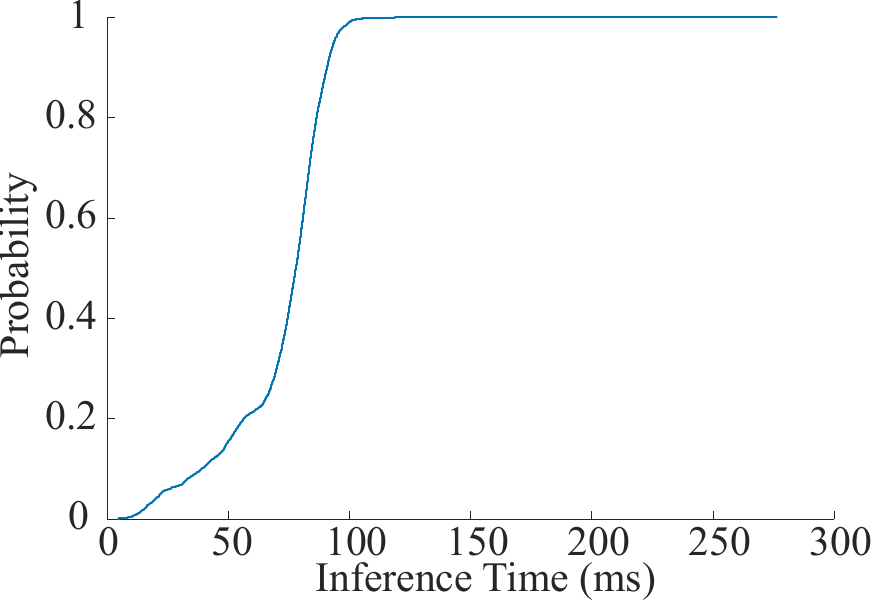}
    \caption{Inference Time.}
    \label{fig:inferencetime}
\end{wrapfigure}
 PALCAS has its own time complexity at each inference round. Figure ~\ref{fig:inferencetime} presents a cumulative distribution function (CDF) of the inference time at each decision-making interval, measured $3000$ times. The result indicates that $90\%$ inference time of an agent for all CAVs is below $100\;\text{ms}$, which is realistic for real-time inference.

\section{Conclusion}
We have presented PALCAS, a priority-guided multi-agent framework for autonomous vehicle lane change decision making based on federated reinforcement learning. PALCAS is equipped with a novel lane-priority advisory system, depending on CAV's routes by considering road safety during lane changes, to effectively navigate vehicles through complex highways with on/off ramps. Our extensive simulations exhibit PALCAS's capabilities of properly aligning vehicles in lanes in congested highways while ensuring safety. A detail analysis of computational time and real-world implementation of PALCAS are our future work.
\bibliographystyle{named}
\bibliography{References/palcas}

\end{document}